\documentclass{article} % For LaTeX2e
\usepackage{iclr2027_conference,times}

\usepackage{amsmath,amsfonts,bm}

\def\eqref#1{equation~\ref{#1}}
\def\1{\bm{1}}

\DeclareMathAlphabet{\mathsfit}{\encodingdefault}{\sfdefault}{m}{sl}
\SetMathAlphabet{\mathsfit}{bold}{\encodingdefault}{\sfdefault}{bx}{n}

\usepackage[utf8]{inputenc}
\usepackage[T1]{fontenc}
\usepackage{url}
\usepackage{booktabs}
\usepackage{amsfonts}
\usepackage{nicefrac}
\usepackage{microtype}
\usepackage{xcolor}
\usepackage{graphicx}
\usepackage{amsmath}
\usepackage{amssymb}
\usepackage{mathtools}
\usepackage{amsthm}
\usepackage{colortbl}
\usepackage{tabularx}
\usepackage{multirow}
\usepackage{wrapfig}
\usepackage{placeins}
\usepackage{hyperref}

\title{One Token Per Frame: Reconsidering Visual Bandwidth in World Models for VLA Policy}

\author{%
  Zuojin Tang$^{1}$\quad
  Shengchao Yuan$^{2}$\quad
  Xiaoxin Bai$^{3}$\quad
  Zhiyuan Jing$^{5}$\quad\\
  \textbf{De Ma$^{1}$}\quad
  \textbf{Gang Pan$^{1*}$}\quad
  \textbf{Bin Liu$^{4*}$} \\
  $^1$Zhejiang University\quad
  $^2$Central South University\quad
  $^3$Harbin Institute of Technology\\
  $^4$Embodied Intelligence General Platform Laboratory, Chery Auto\\
  $^5$E-surfing Digital Life Technology Co., Ltd., China Telecom \\
  $^*$Corresponding Author\\
}

\iclrfinalcopy 
\begin{document}
\maketitle

\lhead{Preprint} % 必须放在 \maketitle 后；用 \lhead{} 可清空页眉

\begin{abstract}
Vision-language-action (VLA) models can use visual prediction to anticipate future states, but dense visual features make the generative sequence grow with the number of camera views, prediction horizon, and encoder resolution. Whether such dense representations are necessary for effective control remains unclear. We introduce \textbf{OneWM-VLA}, which represents each retained camera view with one predictive token per future step. Adaptive Attention Pooling compresses visual features into compact latents, which are jointly generated with robot actions under a conditional flow-matching objective. Future observations provide the latent targets during training and are not required at inference. This design incorporates visual prediction into a pretrained VLA policy while keeping the generative sequence compact. On MetaWorld~MT50, OneWM-VLA improves the average success rate of the $\pi_0$ backbone from $47.91\%$ to $61.53\%$, reaching $72.01\%$ after 60k training steps. It also achieves $98.1\%$ success on LIBERO and raises \emph{Fold Cloth} success on a real Piper arm from $20.0\%$ to $60.0\%$ relative to $\pi_0$. Comparisons on two additional VLA backbones consistently favor one token over three across the evaluated checkpoints. A matched ablation at a longer action horizon further shows that removing the latent loss reduces success from $58.09\%$ to $21.64\%$, supporting the benefit of future supervision for policy learning.
\end{abstract}

\section{Introduction}
\label{sec:intro}

Vision-language-action (VLA) models map visual observations, language instructions, and robot states to actions. Recent systems~\cite{brohan2022rt,zitkovich2023rt,kim2024openvla,kim2025fine,chi2025diffusion,black2026pi0visionlanguageactionflowmodel,intelligence2025pi_} show that large multimodal backbones can support many language-conditioned manipulation tasks. Long-horizon control, however, requires more than reacting to the current observation. A policy must retain task-relevant state, anticipate changes in contacts and object configurations, and keep local errors from accumulating over time. This has motivated VLA policies that augment action generation with predicted frames, latent states, or other future targets~\cite{cen2025worldvla,zhang2025dreamvla,hu2024video}.

Prediction creates a basic design tradeoff. Dense future features preserve spatial detail, but every retained view contributes another block of visual tokens at every future step. When one transformer co-generates predictions and actions, every additional predictive token participates in each sampling step. The sequence length therefore grows with both the prediction horizon and the encoder resolution. This cost is especially relevant when the predictive stream is added to a pretrained VLA under a limited adaptation budget. Recurrent state-space models~\cite{hafner2019dream,hafner2025training} and joint-embedding predictors~\cite{lecun2022path,bardes2024revisiting} suggest that useful future information need not take the form of dense reconstruction, but they do not determine how much visual information a generative VLA should retain for control.

We refer to this allocation as \emph{predictive visual bandwidth}: the number of visual tokens assigned to each retained view and future step. This bandwidth controls both the capacity of the predictive representation and the length of the generated suffix. An overly narrow representation may discard task-relevant changes, whereas a wide representation may increase memory and sampling cost without improving the action policy. This makes the operating point an empirical design choice for the joint visual--action generator, rather than a direct consequence of the encoder patch count. Our central question is whether a single predictive token per view can retain useful future structure while keeping the generated trajectory compact.

\begin{figure*}[t]
  \centering
  \includegraphics[width=\columnwidth]{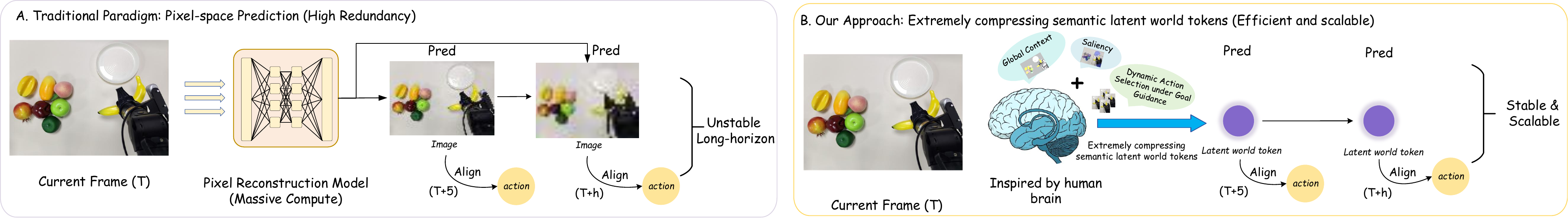}
  \caption{OneWM-VLA represents each retained camera view at each future step with one predictive token. Total suffix length grows with the horizon, while the per-step predictive bandwidth is independent of the encoder patch count.}
  \label{motivation}
\end{figure*}

We introduce \textbf{OneWM-VLA} (Fig.~\ref{motivation}) to study this question. Adaptive Attention Pooling compresses the encoder features of each retained camera view into one token per frame. During training, the same module extracts targets from future frames, producing a compact latent trajectory aligned with the demonstrated actions. A conditional flow then generates the latent and action trajectories together. The interleaved sequence places each future latent block before its action block, allowing the action stream to use the current latent estimate while both trajectories are generated. The latent trajectory remains internal to the policy, while only the generated actions are executed. We initialize the policy from $\pi_0$~\cite{black2026pi0visionlanguageactionflowmodel} and adapt the backbone with LoRA~\cite{hu2022lora} together with the pooling and projection modules.

We evaluate OneWM-VLA at three levels. At the policy level, it improves the average success rate of $\pi_0$ from $47.91\%$ to $61.53\%$ under the primary MetaWorld~MT50 protocol, reaches $98.1\%$ on LIBERO, and, relative to $\pi_0$, raises success on the real-world \emph{Fold Cloth} task from $20.0\%$ to $60.0\%$. At the design level, we study predictive bandwidth, longer training, pooling, target construction, latent supervision, and action horizon, with additional bandwidth comparisons on $\pi_{0.5}$ and SmolVLA. At the representation level, we test whether the latent representation collapses and whether it retains task structure. Together, these experiments evaluate the performance and cost of the compact interface across training durations and backbones.

Our contributions are as follows:
{
\setlength{\leftmargini}{10pt}
\begin{itemize}
    \item We formulate predictive visual bandwidth as a design choice in generative VLA policies and study its relation to sequence length and control performance.
    \item We introduce OneWM-VLA, which compresses each retained view into one predictive token and co-generates the latent and action trajectories in the same flow.
    \item We evaluate the design across three manipulation settings, with longer training, additional VLA backbones, alternative target constructors, and representation diagnostics.
\end{itemize}
}

\section{Related Work}

\subsection{Vision-Language-Action Models}
Progress in multimodal language models~\cite{beyer2024paligemma,li2023vision,wang2024qwen2,bai2025qwen2,bai2025qwen3vltechnicalreport} and large robot datasets has led to VLA policies that map images, instructions, and robot state to actions~\cite{brohan2022rt,zitkovich2023rt,kim2024openvla,intelligence2025pi_,zhao2023learning,chen2025pi_,black2026pi0visionlanguageactionflowmodel,tang2025vlascd,qu2025spatialvla,bu2025univla,zheng2025x,kim2025fine,goyal2025vla,team2024octo,shukor2025smolvla}. Diffusion and flow-matching policies~\cite{chi2025diffusion,black2026pi0visionlanguageactionflowmodel,intelligence2025pi_,liu2024rdt} model multimodal action distributions; other methods add reasoning traces, auxiliary heads, or intermediate multimodal outputs~\cite{zhao2025cot,ye2025vla,zeng2025janusvln,huang2025thinkact,team2025gemini,li2024cogact}. OneWM-VLA retains the existing flow-based action generator and studies how much bandwidth to allocate to its predictive visual stream.

\subsection{Predictive Representations and Latent Planning}
Compact visual representations and learned rewards have been widely studied for robotic control~\cite{nair2022r3m,ma2022vip,ma2023liv,li2024decisionnce}.
Pixel- and video-space methods predict future observations as goals or auxiliary targets~\cite{cen2025worldvla,bjorck2025gr00t,cheang2024gr,zhen20243d,xiao2025world,du2023learning,jang2025dreamgen}, while subgoal generation and latent planning provide future guidance for downstream policies~\cite{xie2025latent,liu2025efficient}; other systems generate visual futures and actions in a shared model~\cite{guo2024prediction,zhang2025up,hu2024video,cen2025rynnvla,jiang2025rynnvla,zhang2025dreamvla}.
Latent state-space models~\cite{hafner2019dream,hafner2025training}, joint-embedding prediction~\cite{lecun2022path,bardes2024revisiting,assran2023self,maes2026leworldmodel,tang2026alam,chen2026abot,tang2026dlam}, and sequence models~\cite{chen2021decision,tang2025efficient} show that useful prediction need not reconstruct every pixel.

Across these lines of work, predictive targets range from dense observations to compact learned states, and joint visual--action generation is already established. OneWM-VLA studies a specific interface within a pretrained flow-based VLA: one predictive token per retained view and future step. Our contribution is the design and empirical evaluation of this compact configuration, including its measured cost and its dependence on target construction, training duration, and backbone.

\begin{figure*}[t]
  \centering
  \includegraphics[width=\columnwidth]{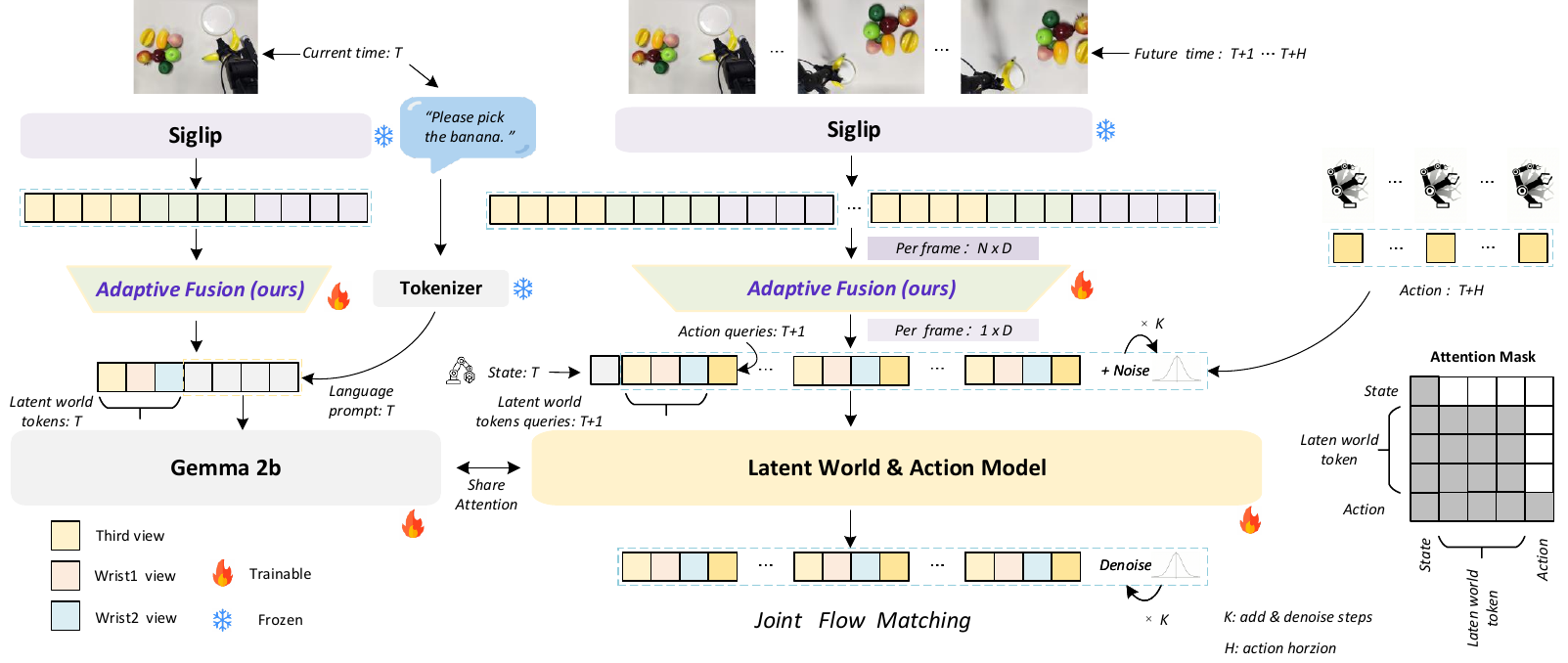}
  \caption{\textbf{OneWM-VLA.} Adaptive Attention Pooling compresses the base and left-wrist views. The expert samples interleaved latent and action blocks with a shared flow time; only the actions are executed.}
  \label{main_framework}
\end{figure*}

\section{Method}
\label{method}
Flow matching learns a velocity field along a path between data and noise, and its conditional form provides a closed-form regression target without ODE integration during training~\cite{lipman2022flow}. OneWM-VLA retains the straight path of the $\pi_0$ action generator and extends it to visual latents. The method has two components (Fig.~\ref{main_framework}). Adaptive Attention Pooling maps each retained view to one latent target token (Sec.~\ref{sec:adaptive_pooling}), and a shared flow generates the future latent and action trajectories together (Sec.~\ref{sec:flow_matching}). We use the $L_1$ velocity-regression loss selected in Appendix~\ref{app:l1_l2}.

\subsection{Per-Frame Compression via Adaptive Attention Pooling}
\label{sec:adaptive_pooling}

% \begin{figure}[t]
%   \centering
%   \includegraphics[width=0.60\linewidth]{onewm_adaptive.pdf}
%   \caption{\textbf{Adaptive Attention Pooling.} Three scoring branches aggregate patch features from each view, and learned branch weights fuse the pooled vectors into one predictive token.}
%   \label{adaptive_fusion}
% \end{figure}

\begin{wrapfigure}{r}{0.5\textwidth}
  \begin{minipage}{0.50\textwidth}
  \begin{center}
    \centerline{\includegraphics[width=1\columnwidth]{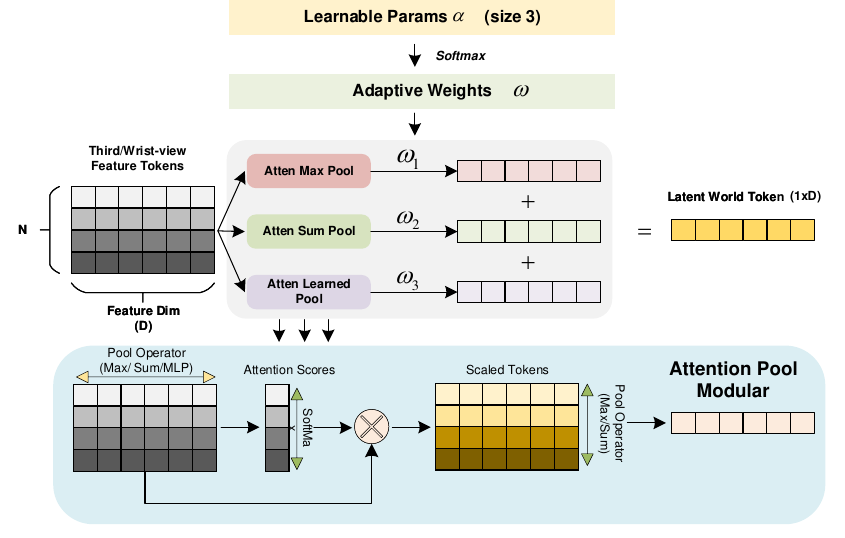}}
     \caption{\textbf{Adaptive Attention Pooling.} Three scoring branches aggregate patch features from each view, and learned branch weights fuse the pooled vectors into one predictive token.}
    \label{adaptive_fusion}
  \end{center}
  \end{minipage}
\end{wrapfigure}

Adaptive Attention Pooling reduces each retained camera view to one token per frame (Fig.~\ref{adaptive_fusion}). For view $i$, the pretrained PaliGemma encoder~\cite{beyer2024paligemma} produces $N{=}256$ patch features, denoted by $\{\mathbf{x}_i^{(n)}\}_{n=1}^{N}$. Each feature lies in $\mathbb{R}^{D}$, with $D{=}2048$. The pooling module scores these features in three complementary ways:
\begin{equation}
\begin{aligned}
\phi_{i,\textsc{Max}}(\mathbf{x}) &= \max_{d}x^{(d)}, &
\phi_{i,\textsc{Sum}}(\mathbf{x}) &= \sum_{d=1}^{D}x^{(d)}, &
\phi_{i,\textsc{Learn}}(\mathbf{x}) &= Q_{\theta_i}(\mathbf{x}).
\end{aligned}
\end{equation}
where $Q_{\theta_i}$ is a small view-specific MLP. Within each branch, a softmax over these scores at temperature $\tau_p$ gives the patch weights. The \textsc{Max} branch applies a coordinate-wise maximum to the weighted features, while the \textsc{Sum} and \textsc{Learn} branches use weighted sums. This produces three pooled vectors $\mathbf{p}_{i,m}$.

Learned logits $\alpha_{i,m}$ define the branch weights that combine these vectors into the final token:
\begin{equation}
\label{eq:pool_fusion}
\mathbf{Z}_i = \sum_{m\in\mathcal{M}}\beta_{i,m}\mathbf{p}_{i,m},
\qquad
\beta_{i,m} = \frac{\exp(\alpha_{i,m}/\tau_p)}
{\sum_{m'\in\mathcal{M}}\exp(\alpha_{i,m'}/\tau_p)},
\end{equation}
where $\mathcal{M}=\{\textsc{Max},\textsc{Sum},\textsc{Learn}\}$. We apply the module independently to the base and left-wrist views, denoted by $\mathcal{V}=\{b,w\}$. The patch weights adapt to each frame, while the branch weights are learned across the training set. The result is one compact token per retained view and frame, independent of the number of encoder patches. Across future steps, these tokens form the clean latent sequences used below. In the direct-$K$ bandwidth comparisons, $K$ counts the fused output tokens actually passed to the model per view and frame; branch fusion remains enabled at every token budget.

\subsection{Joint Latent--Action Flow Matching}
\label{sec:flow_matching}

A classical transition model predicts $p(s_{t+1}\mid s_t,a_t)$. OneWM-VLA instead treats both future actions and visual latents as generated variables and models
\begin{equation}
p_\theta\!\left(a,\{z_i\}_{i\in\mathcal{V}}\mid o_t,\ell,s_t\right).
\end{equation}
This defines a conditional flow over aligned action and latent sequences. The two streams share one flow time and are updated synchronously. At each future step, the action block can use the current latent iterates without waiting for a completed latent rollout. This differs from a two-stage latent planner followed by inverse dynamics: no observed future state or latent is available at inference, and action updates do not wait for the latent trajectory to finish.

During training, the same visual encoder and pooling module are applied to the corresponding future frames to construct clean latent targets; future frames are never inference inputs. Let $a$ denote the future action sequence and $z_i$ the latent sequence for view $i$. These latent variables enter the expert as continuous features. Following the straight path used by $\pi_0$~\cite{black2026pi0visionlanguageactionflowmodel}, the action and latent variables share one sampled flow time:
\begin{equation}
\begin{aligned}
x_t^a &= t\epsilon_a+(1-t)a, &
x_t^{z_i} &= t\epsilon_{z_i}+(1-t)z_i, \\
t &\sim \mathrm{Beta}(1.5,1).
\end{aligned}
\end{equation}
where the noise terms are sampled independently from a standard Gaussian. The target velocities are $u_t^a=\epsilon_a-a$ and $u_t^{z_i}=\epsilon_{z_i}-z_i$.

The pooled current-view tokens and language tokens form the conditioning prefix, followed by the current robot-state block; neither attends to the generated suffix. The suffix interleaves latent and action variables by future step, with the latent queries placed before the action query at each step. Each latent block attends to the conditioning blocks, latent blocks through the same future step, and action blocks at earlier steps. Each action query attends to the conditioning blocks, all earlier suffix blocks, its own noisy iterate, and the latent iterates at the same step. Training and inference use the same block-visibility pattern.

A shared transformer predicts the action and latent velocity fields with the joint objective
\begin{equation}
\mathcal{L}
=\lambda_a\,\mathbb{E}\!\left[\|v_\theta^a-u_t^a\|_1\right]
+\sum_{i\in\mathcal{V}}\lambda_i\,
\mathbb{E}\!\left[\|v_\theta^{z_i}-u_t^{z_i}\|_1\right].
\end{equation}
The latent and action streams share the generator, flow time, and attention layers, so their predictions evolve together during ODE sampling. View-specific linear input and output projections connect each latent stream to the expert, while a separate head predicts action velocity. Thus, the latent and action variables are updated in one conditional flow, while the visibility mask allows each action block to use the corresponding latent iterates.

\paragraph{Future-target supervision.}
The latent velocity loss is the only objective that directly supervises the future-latent stream. Its clean endpoints are extracted from the corresponding observed future frames, tying each latent target to a specific future observation and time step. We examine target variation and task separation through covariance-spectrum and feature-space diagnostics in Sec.~\ref{sec:representation_diagnostics}.

At inference, let $c=(o_t,\ell,s_t)$ denote the observed context and let $\mathbf{x}_t$ collect all action and view-specific latent variables. Starting from $\mathbf{x}_1\sim\mathcal{N}(0,I)$, we solve
\begin{equation}
\frac{\mathrm{d}\mathbf{x}_t}{\mathrm{d}t}
=v_\theta\!\left(t,\mathbf{x}_t\mid c\right)
\end{equation}
from $t{=}1$ to $t{=}0$. Only the action component of $\mathbf{x}_0$ is executed; the generated latent trajectory remains internal to the policy.

\section{Experiments}
\label{sec:experiments}

We evaluate our method on MetaWorld~MT50, LIBERO, and a real Piper arm. MetaWorld is the primary benchmark, with a five-step action horizon used for the primary $\pi_0$ experiments unless stated otherwise. We then organize the remaining evidence as ablations and diagnostics of predictive bandwidth, target construction, latent supervision, representation quality, and temporal robustness.

\begin{figure*}[t]
  \centering
  \includegraphics[width=\columnwidth]{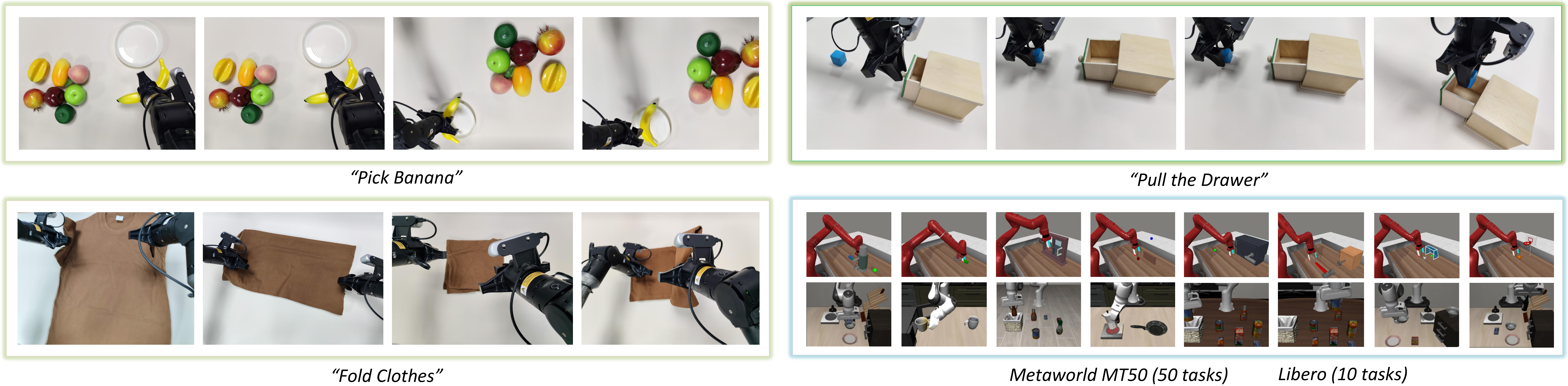}
  \caption{Evaluation suites used in this work: the LIBERO and MetaWorld MT50 simulation benchmarks, and a real Piper robot arm.}
  \label{exp_overview}
\end{figure*}

\subsection{Experimental Setup}
\label{sec:experimental_setup}

\paragraph{Benchmarks.}
\textbf{MetaWorld~MT50}~\cite{mclean2025Metaworld} contains $50$ manipulation tasks, grouped into Easy, Medium, Hard, and Very Hard tiers (Appendix~\ref{Task_definitions}). We evaluate each task for $10$ episodes, average task success rates within each tier, and report the unweighted mean of the four tier rates. This convention applies to all MetaWorld~MT50 average success rates in the main text and appendix. A separate study evaluates longer action horizons of up to $30$ steps. \textbf{LIBERO}~\cite{liu2023libero} contains the Spatial, Object, Goal, and Long suites, with Long emphasizing extended-horizon manipulation. Overall success is the unweighted mean of the four suite success rates. \textbf{Real-world Piper} uses a 6-DoF arm with a parallel gripper. We evaluate \emph{Pick Banana}, \emph{Fold Cloth}, and \emph{Pull Drawer} with $20$ trials per task in each of two conditions: clean observations and combined perturbations involving lighting changes, object shifts, and unseen distractors. For each condition, overall success is the mean of the three task success rates. Training and inference use action chunks of $50$ steps. Figure~\ref{exp_overview} summarizes the three evaluation settings.

\paragraph{Implementation.}
We build OneWM-VLA on $\pi_0$~\cite{black2026pi0visionlanguageactionflowmodel} and use LoRA~\cite{hu2022lora} for adaptation. Dense non-LoRA Gemma weights remain frozen, while SigLIP, pooling, and projection modules are trained along with the LoRA adapters. Sampling uses $10$ ODE steps; the action loss has weight $1.0$, and each latent loss has weight $0.1$. The primary MetaWorld models are trained for 30k steps on $8$ NVIDIA A800 GPUs. For the main LIBERO evaluation, all suites share one checkpoint trained with an action horizon of $20$; inference settings are selected on a held-out validation set (Appendix~\ref{app:unlock_long_horizon}).

\paragraph{Baselines.}
Our primary backbone baseline is $\pi_0$~\cite{black2026pi0visionlanguageactionflowmodel}. The LIBERO and real-robot comparisons also include $\pi_{0.5}$~\cite{intelligence2025pi_}. We also report available results for Diffusion Policy~\cite{chi2025diffusion}, ACT~\cite{zhao2023learning}, Octo~\cite{team2024octo}, OpenVLA~\cite{kim2024openvla}, SpatialVLA~\cite{qu2025spatialvla}, CoT-VLA~\cite{zhao2025cot}, WorldVLA~\cite{cen2025worldvla}, GR00T-N1~\cite{bjorck2025gr00t}, and DreamVLA~\cite{zhang2025dreamvla}. For methods drawn from prior work, we retain the values in their published evaluation records. Paired ablations use a common training and evaluation protocol within each study.

\subsection{Overall Performance}

\begin{table*}[!t]
\centering
\caption{MetaWorld MT50 success rates (\%). $\Delta_{\pi_0}$ is the percentage-point gain over $\pi_0$ at the same horizon. The five-step setting is our primary protocol.}
\label{tab:main_Metaworld_results}
\vspace{2mm}
\footnotesize
\setlength{\tabcolsep}{4pt}
\begin{tabular}{l c c c c c c c}
\toprule
\textbf{Method} & \textbf{Horizon} & \textbf{Easy} & \textbf{Medium} & \textbf{Hard} & \textbf{Very Hard} & \textbf{Avg\%} & \textbf{$\Delta_{\pi_0}$} \\
\midrule
Diff. Policy & 5 & 8.21 & 0.00 & 0.00 & 0.00 & 2.05 & -- \\
ACT & 5 & 31.43 & 11.82 & 0.00 & 4.00 & 11.81 & -- \\
\rowcolor{gray!12} $\pi_0$ & 5 & 71.79 & \textbf{48.18} & 41.67 & 30.00 & 47.91 & -- \\
\rowcolor{blue!8} \textbf{OneWM-VLA} & 5 & \textbf{79.29} & 41.82 & \textbf{65.00} & \textbf{60.00} & \textbf{61.53} & \textbf{+13.62} \\
\midrule
Diff. Policy & 10 & 9.64 & 0.00 & 0.00 & 0.00 & 2.41 & -- \\
ACT & 10 & 47.50 & 7.27 & 0.00 & 4.00 & 14.69 & -- \\
\rowcolor{gray!12} $\pi_0$ & 10 & 77.50 & 30.91 & 30.00 & \textbf{48.00} & 46.60 & -- \\
\rowcolor{blue!8} \textbf{OneWM-VLA} & 10 & \textbf{78.57} & \textbf{31.82} & \textbf{58.33} & 40.00 & \textbf{52.18} & \textbf{+5.58} \\
\midrule
Diff. Policy & 25 & 8.93 & 0.00 & 0.00 & 0.00 & 2.23 & -- \\
ACT & 25 & 31.43 & 0.00 & 0.00 & 0.00 & 7.86 & -- \\
\rowcolor{gray!12} $\pi_0$ & 25 & 73.93 & 27.27 & \textbf{36.67} & 22.00 & 39.97 & -- \\
\rowcolor{blue!8} \textbf{OneWM-VLA} & 25 & \textbf{74.64} & \textbf{30.00} & 25.00 & \textbf{54.00} & \textbf{45.91} & \textbf{+5.94} \\
\midrule
Diff. Policy & 30 & 6.43 & 0.00 & 0.00 & 0.00 & 1.61 & -- \\
ACT & 30 & 31.43 & 0.00 & 0.00 & 0.00 & 7.86 & -- \\
\rowcolor{gray!12} $\pi_0$ & 30 & \textbf{80.00} & \textbf{35.45} & 35.00 & 4.00 & 38.61 & -- \\
\rowcolor{blue!8} \textbf{OneWM-VLA} & 30 & 72.86 & 33.64 & \textbf{45.00} & \textbf{60.00} & \textbf{52.87} & \textbf{+14.26} \\
\bottomrule
\end{tabular}
\end{table*}

\begin{table}[!h]
\centering
\begin{minipage}[t]{0.47\linewidth}
\vspace{0pt}
\caption{Main results on LIBERO. Success rates (\%) on the four suites. Inference-time configuration is detailed in Appendix~\ref{app:unlock_long_horizon}.}
\label{tab:main_result_libero}
\end{minipage}%
\hfill
\begin{minipage}[t]{0.48\linewidth}
\vspace{0pt}
\caption{Real-world Piper manipulation results. Success rates (\%) on three tasks under clean observations and combined perturbations.}
\label{tab:real_world_results}
\end{minipage}

\vspace{1mm}
\noindent
\begin{minipage}[t]{0.47\linewidth}
\vspace{0pt}
\centering
\footnotesize
\setlength{\tabcolsep}{2pt}
\renewcommand{\arraystretch}{1.0}
\begin{tabular}{lccccc}
\toprule
\textbf{Method} & \textbf{Spatial} & \textbf{Object} & \textbf{Goal} & \textbf{Long} & \textbf{Avg\%} \\
\midrule
Diffusion Policy & 78.3 & 92.5 & 68.3 & 50.5 & 72.4 \\
Octo             & 78.9 & 85.7 & 84.6 & 51.1 & 75.1 \\
OpenVLA          & 84.7 & 88.4 & 79.2 & 53.7 & 76.5 \\
SpatialVLA       & 88.2 & 89.9 & 78.6 & 55.5 & 78.1 \\
CoT-VLA          & 87.5 & 91.6 & 87.6 & 69.0 & 83.9 \\
WorldVLA         & 87.6 & 96.2 & 83.4 & 60.0 & 81.8 \\
SmolVLA          & 93.0 & 94.0 & 91.0 & 77.0 & 88.8 \\
GR00T-N1         & 94.4 & 97.6 & 93.0 & 90.6 & 93.9 \\
DreamVLA         & 97.5 & 94.0 & 89.5 & 89.5 & 92.6 \\
\rowcolor{gray!12} $\pi_0$
                  & 96.8 & 98.8 & 95.8 & 85.2 & 94.2 \\
$\pi_{0.5}$       & \textbf{98.8} & 98.2 & 98.0 & 92.4 & 96.9 \\
\rowcolor{blue!8} \textbf{OneWM-VLA}
                  & 98.2 & \textbf{99.6} & \textbf{99.0}
                  & \textbf{95.6} & \textbf{98.1} \\
\bottomrule
\end{tabular}
\end{minipage}%
\hfill
\begin{minipage}[t]{0.48\linewidth}
\vspace{0pt}
\centering
\footnotesize
\setlength{\tabcolsep}{1pt}
\renewcommand{\arraystretch}{1.26}
\begin{tabular}{lcccc}
\toprule
\textbf{Method} & \textbf{P.Banana} & \textbf{F.Cloth} & \textbf{P.Drawer} & \textbf{Avg\%} \\
\midrule
\multicolumn{5}{l}{\textit{Clean conditions}}\\
\midrule
\rowcolor{gray!12} $\pi_0$ & \textbf{100.0} & 20.0 & 30.0 & 50.0 \\
$\pi_{0.5}$& \textbf{100.0} & 25.0 & 50.0 & 58.3 \\
\rowcolor{blue!8} \textbf{OneWM-VLA} & \textbf{100.0} & \textbf{60.0} & \textbf{55.0} & \textbf{71.7} \\
\midrule
\multicolumn{5}{l}{\textit{With observation noise}}\\
\midrule
\rowcolor{gray!12} $\pi_0$ & 65.0 & 0.0 & 10.0 & 25.0 \\
$\pi_{0.5}$ & 60.0 & 10.0 & 25.0 & 31.7 \\
\rowcolor{blue!8} \textbf{OneWM-VLA} & \textbf{75.0} & \textbf{40.0} & \textbf{35.0} & \textbf{50.0} \\
\bottomrule
\end{tabular}
\end{minipage}
\end{table}

\textbf{MetaWorld~MT50.} At five steps, our method attains $61.53\%$ average success, a $13.62$-point gain over $\pi_0$ (Tab.~\ref{tab:main_Metaworld_results}). The gain is concentrated on the Hard and Very Hard tiers; the Medium tier decreases from $48.18\%$ to $41.82\%$, so the aggregate gain does not extend to every task group. The remaining rows use matched non-primary horizons and are analyzed separately in Sec.~\ref{sec:q4}.

\textbf{LIBERO.} OneWM-VLA achieves $98.1\%$ average success, above $\pi_0$ and $\pi_{0.5}$, with its clearest advantage on the Long suite (Tab.~\ref{tab:main_result_libero}). The same checkpoint is used across all four suites; Appendix~\ref{app:unlock_long_horizon} lists the fixed inference settings. The Object and Goal suites are close to saturation for these three models, whereas the Long suite leaves a clearer margin. The larger margin on Long is consistent with the intended use of the design in extended interactions; the comparison evaluates the full policy and its validation-selected inference settings.

\textbf{Real-world Piper.} Relative to $\pi_0$, OneWM-VLA raises the clean-condition average from $50.0\%$ to $71.7\%$, with the largest gain on \emph{Fold Cloth}, where success rises from $20.0\%$ to $60.0\%$ (Tab.~\ref{tab:real_world_results}). Under the combined observation-perturbation setting, it has the highest result on all three tasks. These results provide an initial real-robot validation of the design. The improvement is largest on the deformable-object task, while clean-condition Pick Banana is tied at $100.0\%$ and Pull Drawer shows a smaller gain.

Across the simulation benchmarks and the real robot, the largest observed gains occur on the more demanding settings: the Hard and Very Hard MetaWorld tiers, LIBERO-Long, and \emph{Fold Cloth}. LIBERO Object/Goal and clean-condition Pick Banana are near saturation, so the aggregate improvement is not a uniform shift across all task groups. This pattern motivates the controlled studies below, which examine the compact interface separately from the overall policy comparisons.

\subsection{Ablations and Diagnostics}

We organize the controlled studies around four questions: how much predictive bandwidth is needed, how the one-token target should be constructed, how the future-latent stream behaves, and whether the gains persist across action horizons. We first examine success and measured cost across token budgets, then training checkpoints and transfer across backbones. These studies evaluate the complete compression configurations; the remaining ablations examine target construction and representation quality. Each study is interpreted within its stated evaluation protocol.

\subsubsection{Q1: How Much Predictive Bandwidth Is Needed?}
\label{sec:q1}

\begin{table*}[t]
\centering
\small
\caption{\textbf{Predictive bandwidth ablation} on MetaWorld MT50 ($H=5$, 30k training steps).
$K$ is the number of tokens per view and frame. The $K=1$ result uses the primary $H=5$ setting in Tab.~\ref{tab:main_Metaworld_results}; OOM denotes out of memory.}
\label{tab:token_count_ablation}
\setlength{\tabcolsep}{6pt}
\begin{tabular}{lccccc}
\toprule
$K$ & Total tokens & Trainable (M) & Memory & Latency (ms) & Avg. Success (\%) \\
\midrule
256 (Full)        & $>2{,}500$ & 488.83          & OOM    & --     & -- \\
6                 & 126        & 482.78          & Stable & 52.95  & 59.43 \\
3                 & 90         & 482.76          & Stable & 52.47  & 60.63 \\
\rowcolor{blue!8}
\textbf{1} & \textbf{66}& \textbf{482.73} & \textbf{Stable} & \textbf{44.69} & \textbf{61.53} \\
\bottomrule
\end{tabular}
\end{table*}

We first investigate predictive bandwidth at an action horizon of $H=5$. Here, $K$ denotes the number of tokens retained per view and frame. Table~\ref{tab:token_count_ablation} compares dense encoder features ($K=256$) with compact configurations ($K\in\{1,3,6\}$). All compact variants use direct-$K$ outputs: the three pooling branches remain fused, and $K$ is the number of fused tokens passed to the model. The same compression is used for current observations and future targets, so this study compares complete compression configurations.

\paragraph{Inference feasibility and compute overhead.} 
Preserving dense encoder features ($K=256$) produces a recorded sequence of more than $2{,}500$ tokens and runs out of memory in the measured setup. All three compact configurations run within that memory budget. From $K=1$ to $K=6$, trainable parameters increase by $0.05$\,M, from $482.73$\,M to $482.78$\,M, while measured latency rises by $18.5\%$, from $44.69$\,ms to $52.95$\,ms. The table's sequence counts should be distinguished from the generated suffix alone, whose length is $H(2K+1)$ for two views and one action token per step. Latents and actions are sampled synchronously, rather than through an autoregressive rollout.

\paragraph{Policy grounding performance.} 
In the primary ablation (Tab.~\ref{tab:token_count_ablation}), the one-token configuration reaches $61.53\%$ average success, compared with $60.63\%$ and $59.43\%$ for three and six tokens. Together with its lower measured latency, these results support the compact configuration as a practical operating point.

\begin{table*}[!t]
\centering
\begin{minipage}[t]{0.47\linewidth}
\centering
\caption{Average success rates (\%) across training checkpoints (horizon $5$) on MetaWorld MT50. Bold marks each row's highest value.}
\label{tab:longer_training_v2}
\vspace{1mm}
\footnotesize
\setlength{\tabcolsep}{9pt}
\begin{tabular}{lccc}
\toprule
Checkpoint & $K=1$ & $K=3$ & $K=6$ \\
\midrule
10k & 48.12 & \textbf{48.72} & 35.91 \\
20k & \textbf{60.25} & 49.37 & 58.92 \\
30k & \textbf{61.53} & 60.63 & 59.43 \\
40k & 58.10 & 58.07 & \textbf{64.80} \\
50k & \textbf{68.31} & 64.20 & 64.05 \\
60k & \textbf{72.01} & 62.03 & 61.49 \\
\bottomrule
\end{tabular}
\end{minipage}%
\hfill
\begin{minipage}[t]{0.49\linewidth}
\centering
\caption{OneWM performance across VLA backbones on MetaWorld MT50 at horizon $5$: average success rates (\%) and paired gains.}
\label{tab:cross_backbone_v2}
\vspace{1mm}
\footnotesize
\setlength{\tabcolsep}{5pt}
\begin{tabular}{llccc}
\toprule
Backbone & Steps & $K=1$ & $K=3$ & Difference \\
\midrule
$\pi_{0.5}$ & 10k & \textbf{62.4} & 60.0 & +2.4 \\
 & 20k & \textbf{65.4} & 65.0 & +0.4 \\
 & 30k & \textbf{70.6} & 68.2 & +2.4 \\
SmolVLA & 10k & \textbf{66.2} & 61.4 & +4.8 \\
 & 20k & \textbf{71.6} & 68.2 & +3.4 \\
 & 30k & \textbf{74.8} & 69.6 & +5.2 \\
\bottomrule
\end{tabular}
\end{minipage}
\end{table*}

\paragraph{Training checkpoints.}
Table~\ref{tab:longer_training_v2} compares $K\in\{1,3,6\}$ under the same training and evaluation settings. At 60k, success is $72.01\%$, $62.03\%$, and $61.49\%$, respectively: one token is higher by $9.98$ points than the best wider configuration at this checkpoint. It is highest at four of the six reported checkpoints, with $K=3$ higher at 10k and $K=6$ higher at 40k. These results support the compact configuration at the longer training budget, while showing that its advantage varies during training.

\paragraph{Transfer across VLA backbones.}
We also integrate OneWM into $\pi_{0.5}$ and SmolVLA-0.45B. One token is higher than three in all six paired backbone/checkpoint comparisons (Tab.~\ref{tab:cross_backbone_v2}), with margins of $2.4$ and $5.2$ points at 30k on the two backbones. This extends the observed ordering beyond $\pi_0$, although the smallest margin is only $0.4$ points. Both variants include OneWM, so these comparisons measure the relative performance of two compression configurations after integration.

\subsubsection{Q2: How Should the Predictive Target Be Constructed?}
\label{sec:q2}

\begin{table*}[!t]
\centering
\begin{minipage}[t]{0.49\textwidth}
\centering
\caption{Pooling ablation on LIBERO at horizon $5$ under the common 30k-step adaptation schedule.}

\label{tab:libero_ablation}
\vspace{1mm}
\footnotesize
\setlength{\tabcolsep}{1.6pt}
% Four effective rows are expanded to match the five-row branch table.
\renewcommand{\arraystretch}{1.25}
\begin{tabular*}{\linewidth}{@{\extracolsep{\fill}}lccccc}
\toprule
\textbf{Method} & \textbf{Spatial} & \textbf{Object} & \textbf{Goal} & \textbf{Long} & \textbf{Avg\%} \\
\midrule
\rowcolor{blue!8} \textbf{OneWM-VLA} & \textbf{96.6} & \textbf{96.8} & \textbf{94.6} & \textbf{85.0} & \textbf{93.3} \\
Static avg-pooling & 77.8 & 78.0 & 94.0 & 41.4 & 72.8 \\
No branch fusion & 72.4 & 70.4 & 43.0 & 4.4 & 47.6 \\
\bottomrule
\end{tabular*}
\end{minipage}%
\hfill
\begin{minipage}[t]{0.49\textwidth}
\centering
\caption{Pooling branch ablation on MetaWorld MT50 (at horizon 5, 30k steps).}

\label{tab:ablation_pooling}
\vspace{1mm}
\footnotesize
\setlength{\tabcolsep}{1.4pt}
\renewcommand{\arraystretch}{1.0}
\begin{tabular*}{\linewidth}{@{\extracolsep{\fill}}lccccc}
\toprule
\textbf{Method} & \textbf{Easy} & \textbf{Medium} & \textbf{Hard} & \textbf{V-Hard} & \textbf{Avg\%} \\
\midrule
\rowcolor{blue!8} \textbf{OneWM-VLA} & 79.29 & \textbf{41.82} & \textbf{65.00} & 60.00 & \textbf{61.53} \\
Only \textsc{Learn} & 78.57 & 29.09 & 35.00 & 60.00 & 50.67 \\
Only \textsc{Max}   & 53.21 & 16.36 & 0.00  & 20.00 & 22.39 \\
Only \textsc{Sum}   & \textbf{86.43} & 26.36 & 46.67 & \textbf{62.00} & 55.36 \\
\bottomrule
\end{tabular*}
\end{minipage}
\end{table*}

We next examine how the encoder features are aggregated into predictive targets. The controlled studies compare the full adaptive multi-branch module with static pooling, removing fusion, and retaining one scoring branch. The LIBERO ablation uses a separate run with $H=5$ and 30k training steps, so its absolute values should not be compared directly with the main checkpoint in Tab.~\ref{tab:main_result_libero}. The full pooling module has the highest aggregate in both studies. On LIBERO, it reaches $93.3\%$, compared with $72.8\%$ for static averaging, and removing branch fusion is substantially worse; the largest separation appears on Long (Tab.~\ref{tab:libero_ablation}). On MetaWorld, the fused module reaches $61.53\%$, above every individual branch (Tab.~\ref{tab:ablation_pooling}). The \textsc{Sum} branch is stronger on two individual tiers, so the aggregate advantage does not imply that the fused module is best on every tier.

\begin{table}[!t]
\centering
\caption{Adaptive pooling and sequence-to-one MLP targets across action horizons. Entries are MetaWorld average success rates; the last column averages the six horizons.}
\label{tab:constructor_horizon_v2}
\small
\setlength{\tabcolsep}{4pt}
\begin{tabular}{lccccccc}
\toprule
Constructor & $H=5$ & $10$ & $15$ & $20$ & $25$ & $30$ & Avg. Success (\%) \\
\midrule
MLP & 58.91 & \textbf{52.36} & 40.18 & 34.91 & 35.27 & 35.85 & 42.91 \\
\rowcolor{blue!8} Adaptive pooling & \textbf{61.53} & 52.18 & \textbf{52.41} & \textbf{58.09} & \textbf{45.91} & \textbf{52.87} & \textbf{53.83} \\
\bottomrule
\end{tabular}
\end{table}

\paragraph{Target constructor at fixed bandwidth.}
We compare Adaptive Attention Pooling with a sequence-to-one MLP using identical encoder features and one-token, $2048$-D targets, with the downstream flow, attention mask, training schedule, and evaluator held fixed. Adaptive pooling achieves higher success at five of the six action horizons in Tab.~\ref{tab:constructor_horizon_v2}, averaging $53.83\%$ compared with $42.91\%$ for the MLP, a gain of $10.92$ percentage points. Over $H\in\{15,20,25,30\}$, the mean gain is $15.77$ points; the MLP is higher at $H=10$ by $0.18$ points. These results support Adaptive Attention Pooling as the target constructor across the tested action horizons.

\subsubsection{Q3: What Does Future Supervision Contribute?}
\label{sec:q3}
\label{sec:representation_diagnostics}

We first compare the full model with an architecture-matched model trained without the latent loss, using 30k training steps and $H=20$ (Tab.~\ref{tab:latent_loss_ablation}). The architecture, training budget, horizon, camera settings, and evaluation protocol are identical; only the latent loss is removed. Average success falls from $58.09\%$ to $21.64\%$, a $36.45$-point decrease, with lower success in every difficulty group. This matched comparison supports the contribution of future-frame supervision to control.

\begin{wraptable}{R}{0.50\columnwidth}
\vspace{-6pt}
\small
\centering
\caption{Task separation on LIBERO-Long with normalized within-task scatter.}
\label{tab:pca_fisher}
\vspace{1mm}
\footnotesize
\setlength{\tabcolsep}{6pt}
\renewcommand{\arraystretch}{1.0}
\begin{tabular}{lccc}
\toprule
\textbf{Representation}
& \shortstack{$\mathrm{tr}(S_B)$\\\textbf{(norm.)}}
& \shortstack{$\mathrm{tr}(S_W)$\\\textbf{(norm.)}}
& \textbf{$F$} \\
\midrule
Before pooling
& 0.524
& 1.000
& 0.524 \\
\rowcolor{blue!8} After pooling
& 0.405
& 1.000
& 0.405 \\
\bottomrule
\end{tabular}
\end{wraptable}

We then examine the variation retained by the compact representation. Appendix~\ref{app:representation_diagnostics} details the representation diagnostics reported in Tab.~\ref{tab:collapse_diagnostics}. At 30k steps, the $H=5$ full model has a target entropy effective rank of $21.42$ and a participation rank of $8.88$, compared with $1.83$ and $1.31$ without the latent loss. The full model retains variation along multiple directions, while the no-loss control is close to rank one. We also compare the full-dimensional Fisher ratio before and after pooling on the same time-aligned LIBERO-Long samples (Appendix~\ref{sec:pca}). The pooled representation retains $77.3\%$ of the reference Fisher ratio (Tab.~\ref{tab:pca_fisher}), decreasing from $0.524$ to $0.405$. Compression therefore weakens task separation while preserving measurable between-task variation relative to within-task variation. This measures retained task variation, rather than the preservation of every control-relevant factor.

\subsubsection{Q4: Do the Gains Persist Across Action Horizons?}
\label{sec:q4}

We next examine the non-primary horizons in Tab.~\ref{tab:main_Metaworld_results}. OneWM-VLA has higher aggregate success than $\pi_0$ at every matched horizon, but the margin is not monotonic. At the longest horizon, the gain is concentrated on Hard and Very Hard tasks, while $\pi_0$ remains stronger on Easy and Medium tasks. These results show that the observed advantages extend across the evaluated action-chunk lengths, without increasing monotonically with horizon.

\section{Conclusion and Discussion}
\label{sec:conclusion_and_discussion}

We presented OneWM-VLA, which represents each retained camera view with one predictive token per future step and generates predictive latents and actions through a joint conditional flow. This design introduces future-frame supervision without carrying dense visual features through the generative sequence. Across MetaWorld~MT50, LIBERO, and a real Piper arm, it achieves higher aggregate success than $\pi_0$, with the clearest gains on difficult and extended interactions. The compact configuration has the highest success at the final checkpoint of the training comparison and outperforms the wider alternative in the paired studies on two additional VLA backbones. The ablations also highlight the roles of future-frame supervision, adaptive pooling, and future-target construction, while the representation analyses show nonconstant targets and retained task-related variation. Overall, our results support compact predictive latents as an effective interface between visual prediction and action generation in the evaluated settings.

\paragraph{Limitations.}
Our results support one token as an effective configuration rather than a universal optimum. The bandwidth sweep changes current-observation and future-target token counts together. The collapse diagnostics are empirical and provide no general guarantee, which would require additional assumptions or explicit objective and architectural constraints. The real-robot evaluation covers only a small set of tasks; broader studies across tasks and robots, and controls that isolate future-token count, remain future work.

% \section*{Statement on the Use of AI}
% We used ChatGPT to assist with manuscript editing and consistency checks of the reported results and their interpretation. No new experimental results were generated through this assistance. The authors are responsible for the final manuscript.

\bibliography{iclr2027_conference}
\bibliographystyle{iclr2027_conference}

\clearpage
\appendix

\section{MetaWorld MT50 Difficulty Tiers}
\label{Task_definitions}

Following~\cite{seo2023masked}, we divide the 50 MetaWorld MT50 tasks into four tiers by manipulation difficulty (Tab.~\ref{tab:task_difficulty}). We use this partition for every tier-level result in the main text. The checkpoint sweep in Tab.~\ref{tab:longer_training_v2} uses camera $x=0.75$ for all three token configurations.

\begin{table*}[!h]
\centering
\caption{MetaWorld MT50 task-difficulty partition used in our evaluation, following~\cite{seo2023masked}.}
\label{tab:task_difficulty}
\vspace{1mm}
\footnotesize
\begin{tabular}{p{2.2cm} p{10.5cm}}
\toprule
\textbf{Difficulty} & \textbf{Tasks} \\
\midrule
Easy (28) & \texttt{button-press}, \texttt{button-press-topdown}, \texttt{button-press-topdown-wall}, \texttt{button-press-wall}, \texttt{coffee-button}, \texttt{dial-turn}, \texttt{door-close}, \texttt{door-lock}, \texttt{door-open}, \texttt{door-unlock}, \texttt{drawer-close}, \texttt{drawer-open}, \texttt{faucet-close}, \texttt{faucet-open}, \texttt{handle-press}, \texttt{handle-press-side}, \texttt{handle-pull}, \texttt{handle-pull-side}, \texttt{lever-pull}, \texttt{plate-slide}, \texttt{plate-slide-back}, \texttt{plate-slide-back-side}, \texttt{plate-slide-side}, \texttt{reach}, \texttt{reach-wall}, \texttt{window-close}, \texttt{window-open}, \texttt{peg-unplug-side} \\
\midrule
Medium (11) & \texttt{basketball}, \texttt{bin-picking}, \texttt{box-close}, \texttt{coffee-pull}, \texttt{coffee-push}, \texttt{hammer}, \texttt{peg-insert-side}, \texttt{push-wall}, \texttt{soccer}, \texttt{sweep}, \texttt{sweep-into} \\
\midrule
Hard (6) & \texttt{assembly}, \texttt{hand-insert}, \texttt{pick-out-of-hole}, \texttt{pick-place}, \texttt{push}, \texttt{push-back} \\
\midrule
Very Hard (5) & \texttt{shelf-place}, \texttt{disassemble}, \texttt{stick-pull}, \texttt{stick-push}, \texttt{pick-place-wall} \\
\bottomrule
\end{tabular}
\end{table*}

\section{Sensitivity to the Fusion Temperature}
\label{app:temperature}

We vary the pooling temperature $\tau_p$ used for token scoring and branch fusion (Sec.~\ref{sec:adaptive_pooling}). Among the three tested values, $\tau_p{=}0.1$ gives the highest average LIBERO success (Tab.~\ref{tab:temperature}). We therefore use it in the main experiments.

\begin{table*}[!h]
\centering
\caption{LIBERO success rates (\%) for different pooling temperatures $\tau_p$.}
\label{tab:temperature}
\begin{tabular}{lccccc}
\toprule
$\tau_p$ & Spatial & Object & Goal & Long & Avg\% \\
\midrule
1.0 & 96.6 & \textbf{97.8} & 89.2 & 77.4 & 90.3 \\
0.5 & \textbf{97.4} & 97.2 & 87.0 & 78.6 & 90.1 \\
\rowcolor{blue!8} 0.1 & 96.6 & 96.8 & \textbf{94.6} & \textbf{85.0} & \textbf{93.3} \\
\bottomrule
\end{tabular}
\end{table*}

\section{Flow-Matching Distance Metric}
\label{app:l1_l2}

We compare $L_1$ and $L_2$ distances for velocity regression in the action and latent branches, holding all other settings fixed (Tab.~\ref{tab:loss_metric_libero}). The $L_1/L_1$ configuration gives the highest average success among the three variants. We use it in the main experiments as an empirical velocity-regression choice; it does not inherit the conditional-mean guarantee associated with squared-error flow matching.

\begin{table*}[!h]
\centering
\caption{LIBERO success rates (\%) for different distance metrics in the action and latent flow-matching losses. Training and inference both use $H{=}5$.}
\label{tab:loss_metric_libero}
\begin{tabular}{lccccc}
\toprule
Distance (Action / Latent) & Spatial & Object & Goal & Long & Avg\% \\
\midrule
$L_2$ / $L_2$ & 94.8 & 94.6 & 87.0 & 74.2 & 87.7 \\
$L_1$ / $L_2$ & 96.4 & \textbf{98.0} & 88.2 & 78.2 & 90.2 \\
\rowcolor{blue!8} $L_1$ / $L_1$ & \textbf{96.6} & 96.8 & \textbf{94.6} & \textbf{85.0} & \textbf{93.3} \\
\bottomrule
\end{tabular}
\end{table*}

\section{LIBERO Inference-Time Configuration}
\label{app:unlock_long_horizon}

We use one checkpoint across all four LIBERO suites and select the inference action horizon and replan step on a held-out validation subset. Tab.~\ref{tab:libero_infer_config} lists the fixed configurations used for evaluation. We also sweep these two settings on LIBERO-Long with the same checkpoint (Tab.~\ref{tab:libero_sweep}). The gray row is the reference configuration of the same checkpoint, and the blue row is the highest-scoring setting in the sweep. The $20/16$ setting already exceeds the $92.4\%$ result of $\pi_{0.5}$, while $20/20$ is the longest inference horizon. Success ranges from $93.4\%$ to $95.6\%$ across the tested grid, while the representation stays fixed at one token per view and frame. For each suite, the selected pair is fixed for all evaluation rollouts and is not adapted per episode.

\begin{table}[!h]
\centering
\caption{Inference-time configuration of OneWM-VLA on LIBERO. All suites use the checkpoint trained with an action horizon of $20$; the inference horizon and replan step are selected per suite from a small validation grid.}
\label{tab:libero_infer_config}
\vspace{1mm}
\small
\setlength{\tabcolsep}{6pt}
\renewcommand{\arraystretch}{1.05}
\begin{tabular}{lccc}
\toprule
\textbf{Suite} & \textbf{Inference horizon} & \textbf{Replan step} & \textbf{Avg. Success (\%)} \\
\midrule
Spatial & 14 & 10 & 98.2 \\
Object  & 14 & 10 & 99.6 \\
Goal    & 18 & 7  & 99.0 \\
Long    & 18 & 12 & 95.6 \\
\bottomrule
\end{tabular}
\end{table}

\begin{table}[!t]
\centering
\caption{LIBERO-Long success rates (\%) across inference action horizons and replan steps using a common checkpoint trained with an action horizon of $20$.}
\label{tab:libero_sweep}
\setlength{\tabcolsep}{3pt}
\begin{tabular}{lrrr}
\toprule
Train/Infer. & Replan & Suffix tokens & Avg. Success (\%) \\
\midrule
\rowcolor{gray!12} 20/15 (baseline) & 10 & 45 & 93.4 \\
20/16 & 10 & 48 & 94.0 \\
20/18 & 10 & 54 & 94.8 \\
20/18 & 11 & 54 & 95.4 \\
\rowcolor{blue!8} 20/18 & 12 & 54 & \textbf{95.6} \\
20/20 & 10 & 60 & 94.0 \\
\bottomrule
\end{tabular}
\end{table}

\section{Latent-Loss Ablation on MetaWorld MT50}
\label{app:latent_loss_ablation}

Table~\ref{tab:latent_loss_ablation} compares the full OneWM-VLA model and its architecture-matched no-latent-loss control after 30k training steps, with $H=20$ at training and inference. Both use the same camera settings and evaluation protocol from Sec.~\ref{sec:experimental_setup}. The control retains the latent tokens, projections, attention mask, and action objective; only the latent-loss weight is set to zero. The full model reaches $58.09\%$ average success, compared with $21.64\%$ for the control, a difference of $36.45$ percentage points.

\begin{table*}[!t]
\centering
\caption{Latent-loss ablation on MetaWorld MT50 ($H=20$, 30k training steps). $\Delta_{\mathrm{full}}$ is the percentage-point difference from the full model.}
\label{tab:latent_loss_ablation}
\vspace{1mm}
\small
\setlength{\tabcolsep}{5pt}
\renewcommand{\arraystretch}{1.05}
\begin{tabular}{lcccccc}
\toprule
\textbf{Method} & \textbf{Easy} & \textbf{Med.} & \textbf{Hard} & \textbf{V-Hard} & \textbf{Avg\%} & $\Delta_{\mathrm{full}}$ \\
\midrule
\rowcolor{blue!8} \textbf{OneWM-VLA} & \textbf{79.64} & \textbf{42.73} & \textbf{60.00} & \textbf{50.00} & \textbf{58.09} & -- \\
No latent loss & 45.36 & 14.55 & 6.67 & 20.00 & 21.64 & $-36.45$ \\
\bottomrule
\end{tabular}
\end{table*}

\section{Representation-Diagnostic Protocol}
\label{app:representation_diagnostics}

Table~\ref{tab:collapse_diagnostics} reports the target-spectrum and matching diagnostics. Both effective ranks are computed from the covariance of future-frame target vectors.

\begin{table*}[!t]
\centering
\caption{Representation diagnostics for the full one-token model and its architecture-matched no-latent-loss control, both trained with $H=5$ for 30k steps. The last row reports the separate architecture-matched policy evaluation at $H=20$ after 30k training steps.}
\label{tab:collapse_diagnostics}
\vspace{1mm}
\small
\setlength{\tabcolsep}{4.6pt}
\renewcommand{\arraystretch}{1.05}
\begin{tabular}{lcc}
\toprule
\textbf{Diagnostic} & \textbf{Full one-token, 30k} & \textbf{No latent loss, 30k} \\
\midrule
Target entropy effective rank & \textbf{21.42} & 1.83 \\
Covariance participation rank & \textbf{8.88} & 1.31 \\
Temporal discrimination & \textbf{77.09\%} & 50.04\% \\
Cross-task discrimination & \textbf{73.52\%} & 49.18\% \\
Global R@1 & \textbf{0.510\%} & 0.051\% \\
Global chance & 0.051\% & 0.051\% \\
Within-task R@1 & \textbf{7.53\%} & 2.43\% \\
Within-task chance & 2.50\% & 2.50\% \\
\midrule
Policy avg. success ($H=20$) & \textbf{58.09\%} & 21.64\% \\
\bottomrule
\end{tabular}
\end{table*}

\paragraph{Checkpoints and samples.}
We evaluate the full one-token model and its architecture-matched no-latent-loss control, both trained with $H=5$ for 30k steps. We use valid base-view records from $49$ MetaWorld tasks and $245$ episodes, yielding $1{,}960$ contexts. Five future steps per context produce $M_z{=}9{,}800$ target vectors. Both models use the same records and covariance calculation.

\paragraph{Spectral metrics.}
Let $\widetilde{Z}\in\mathbb{R}^{M_z\times D}$ contain the centered target vectors. We compute
\begin{equation}
\label{eq:representation_spectrum}
\begin{aligned}
C_z &= \frac{1}{M_z-1}\widetilde{Z}^{\top}\widetilde{Z},
& p_j &= \frac{\lambda_j}{\sum_k\lambda_k},\\
r_{\mathrm{ent}} &= e^{-\sum_j p_j\log p_j},
& r_{\mathrm{part}} &= \frac{1}{\sum_j p_j^2},
\end{aligned}
\end{equation}
where $\{\lambda_j\}$ are the eigenvalues of $C_z$. Zero covariance corresponds to a constant representation; for nonzero covariance, both effective ranks approach one as the variance concentrates in a single direction.

\paragraph{Matching diagnostics.}
Queries are the generated $2048$-D base-view latents for the five consecutive future frames $t+1,\ldots,t+5$, jointly sampled with actions conditioned on the current image, robot state, and language instruction under a common Gaussian initialization.
For R@1, we subtract the corresponding global or within-task target mean from both queries and pooled future-frame targets, apply $\ell_2$ normalization, and rank targets at the same future offset by cosine similarity. The candidate sets contain $1{,}960$ global targets or $40$ same-task targets, giving chance levels of $0.051\%$ and $2.50\%$, respectively.
Pairwise discrimination measures how often the matched target has a lower target-variance-normalized MSE than a negative, with a $50\%$ chance level. Temporal negatives are constructed by cyclically shifting the eight anchors within each episode by four anchor positions; cross-task negatives are constructed by rotating task identities while preserving the episode slot, anchor slot, and future offset.

\paragraph{Interpretation.}
The full model has target variance distributed across multiple directions, while the no-loss control is close to rank one. The separate $H=20$ policy evaluation at 30k also favors the full model by $36.45$ percentage points (Tab.~\ref{tab:latent_loss_ablation}). The spectral measurements describe representation variation, while the policy evaluation measures the effect of removing the latent loss on task success.

\section{Feature-Space Separation Analysis}
\label{sec:pca}

\paragraph{Features.}
We extract visual features only, without language or action tokens. Before pooling, each sample contains the $256$ visual tokens produced by the PaliGemma encoder. We average these tokens to obtain one feature vector per sample, so the reference is mean-pooled encoder features rather than the full patch sequence. After pooling, each sample is represented directly by the single output token of Adaptive Attention Pooling. This construction gives one full-dimensional feature vector per sample in both settings.

\paragraph{Sampling.}
We use $24$ successful trajectories from each of the $10$ LIBERO-Long tasks. One frame is extracted from each trajectory at the same relative time, yielding $240$ samples in total. The same samples are used before and after pooling.

\paragraph{Fisher ratio.}
Let $\mathcal{I}_c$ contain the samples from task $c$, let $M_c=|\mathcal{I}_c|$, and let $\boldsymbol{\mu}_c$ and $\boldsymbol{\mu}$ denote the task and global feature means. We compute task separation in the original feature space as
\begin{equation}
F
=
\frac{
\sum_c M_c
\left\lVert
\boldsymbol{\mu}_c-\boldsymbol{\mu}
\right\rVert_2^2
}{
\sum_c\sum_{n\in\mathcal{I}_c}
\left\lVert
\mathbf{f}_n-\boldsymbol{\mu}_c
\right\rVert_2^2
}
=
\frac{\mathrm{tr}(S_B)}
{\mathrm{tr}(S_W)},
\end{equation}
where $S_B$ and $S_W$ are the between-task and within-task scatter matrices. The metric is computed from the full feature vectors rather than a low-dimensional projection.

Because the two representations have different feature scales, we normalize both scatter traces by $\mathrm{tr}(S_W)$ within each setting when reporting Tab.~\ref{tab:pca_fisher}. This sets the normalized within-task trace to one and leaves the Fisher ratio unchanged. The resulting comparison measures the relative change in task separation, rather than differences in raw feature magnitude.

\end{document}